# MULTI-USER AUGMENTED REALITY APPLICATION FOR VIDEO COMMUNICATION IN VIRTUAL SPACE

*Kumar Mridul, M. Ramanathan, Kunal Ahirwar, Mansi Sharma*

Indian Institute of Technology Madras, Chennai - 600036

**ABSTRACT**

Communication is the most useful tool to impart knowledge, understand ideas, clarify thoughts and expressions, organize plan and manage every single day-to-day activity. Although there are different modes of communication, physical barrier always affects the clarity of the message due to the absence of body language and facial expressions. These barriers are overcome by video calling, which is technically the most advance mode of communication at present. The proposed work concentrates around the concept of video calling in a more natural and seamless way using Augmented Reality (AR). AR can be helpful in giving the users an experience of physical presence in each other's environment. Our work provides an entirely new platform for video calling, wherein the users can enjoy the privilege of their own virtual space to interact with the individual's environment. Moreover, there is no limitation of sharing the same screen space. Any number of participants can be accommodated over a single conference without having to compromise the screen size.

***Index Terms*** — Virtual/Augmented reality, video call, realistic tele-immersion, remote collaboration, Mobile AR

## 1. BACKGROUND AND OBJECTIVES

The smartphones have become a commonly used gadget to serve multiple purposes in the present time. With the advent of novel technologies and applications, life has become simpler and accelerated. In this fast paced world, being physically present in every situation is not possible. This potentially increases demand of novel tools for Mobile video calling and conferencing. Video conferencing with a smartphone could be used to communicate across any part of the world.

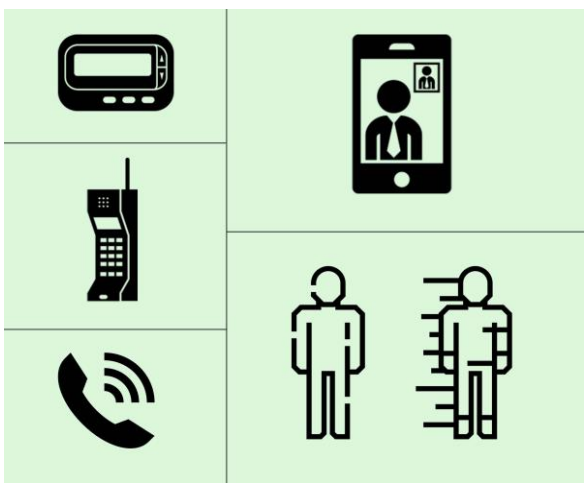

**Figure 1: Evolution of Communication Devices**

Video conferencing helps multiple users to connect over a common platform so as to interact with each other with the aid of audio and video. The major issue arises when there are too many people joining the same platform. It becomes quite difficult to accommodate too many users over the same call because screen dimensions are fixed. More the number of users that join a call, the smaller becomes their window of display, thereby making it very frustrating for every user to even be merely visible. The display size of each user does not only become smaller with every increasing number of participants, but also deviates the purpose of such a conference call. The entire call is bewildered and the idea is not even communicated at its required levels.

Our objective is to create a communication channel in AR [2, 3] wherein many participants can be placed all around the real space of the environment, bringing each participant a step closer to in-place, face-to-face communication compared to a normal video calling application. Such immersion will bring life to the communication and give an essence of natural one on one interaction.

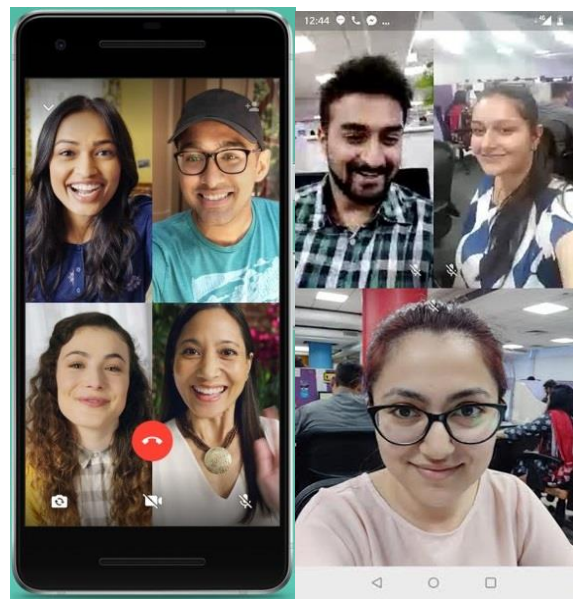

**Figure 2: Current video conferencing solution**

## 2. METHOD

To implement this project, we are using Unity Game Engine [9] with different plugins. There are three main factors that need to be taken care of:

1 – Communication
2 – Augmented Reality

3 – Bringing AR video call on Mobile device

**Communication**

The main idea of this project is to establish a connection among different devices. We used plugins provided by agora.io [8] to establish the connection. The plugin takes care of communication channel, connection and transfer of audio and video data in real-time. The implementation is done in Unity to establish the communication channel.

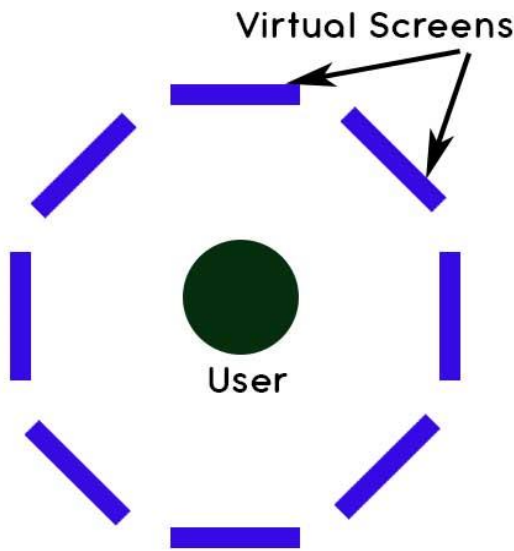

**Figure 3: Virtual screens in users' reality**

**Augmented Reality**

Augmented Reality is an upcoming technology, which lets users visualize virtual objects into the real world. In a typical AR set-up, the real world is captured as it is and virtual information is overlaid upon the real world, making sure that virtual object does not move with respect to the real world when the user moves. This lets the user feel that the virtual world is a part of the real world, thereby creating augmentation. Environment tracking, light estimation, motion, occlusion are major components of AR implementation.

The major role of Augmented Reality has been achieved in the proposed work with the aid of Unity and ARCore [11]. This plugin helps us to develop other Augmented Reality applications. It recognizes the feature points and merges this information with IMU (Inertial Measurement Unit). A tightly coupled algorithm is written to utilize the camera and IMU information. It is very efficient on the devices that support ARCore. The surfaces or walls can be created and mapped by showing them using the phone screen.

**Bringing on Mobile Device**

Mobile devices have been successfully utilizing a lot of applications for enhancing communication. Most of these applications end up giving the same or at least similar features, but nothing in addition to the existing ones. So, bringing a revolution in the mode of video communication will be highly rewarding. However, introducing these traits in the application and using them on the devices is actually challenging.

Bringing the application on the mobile devices [7] is quite cumbersome and tough. This application runs only over selected devices or smartphones that are supported by ARCore [12]. These devices include all the Google Pixel, Nokia 6.1, Nokia 7, etc. The application requires a special permission to be granted. The permission seeks for using both, the primary and the secondary camera of the phone. Once the .apk is created and installed on the phone, the channel (just like a chat room) can be joined. This room is open and active for other users to join the same channel. Once the participant joins, their screen adds up in the user's reality.

## 3. RESULTS

We created an application using the proposed methodology which runs on the phone and gives the recommended results. As it is a crude model, it requires some modification. We are working on UI of the application. As per now, we can add 16 people in the conference Video AR call.

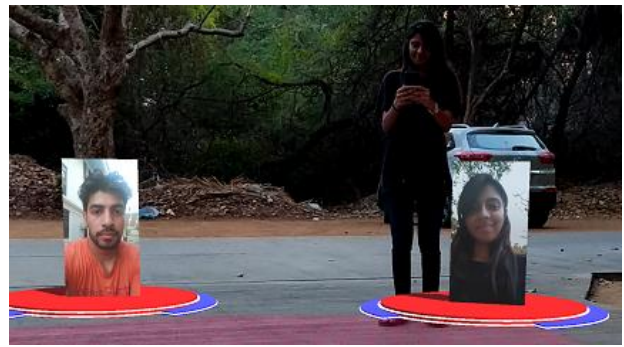

**Figure 4: Proposed Video Conference in AR**

## 4. NOVELTY

Video calling and conferencing mitigate the physical barriers. However, in the modern era it is not highly competent to meet the ever increasing needs to the transfer of knowledge and manpower. Augmented Reality being state-of-the-art technology is effectively used in video conferencing. Using AR in communication adds a new dimension to the current solution. The major problem of the windows stacking over increasing number of participants of the call will be effectively reduced using this application. Moreover, adding many participants on a conference is not possible. There are particular restrictions over the number of users joining a call. The proposed novel model solves this problem by providing space for accommodating the windows of many participants in the user's real environment. It also enables a more realistic presence of the communicators while providing a mode for effective and efficient communication, without causing any chaos and confusion for interaction. A demo of the proposed model is available (see the link [13]).

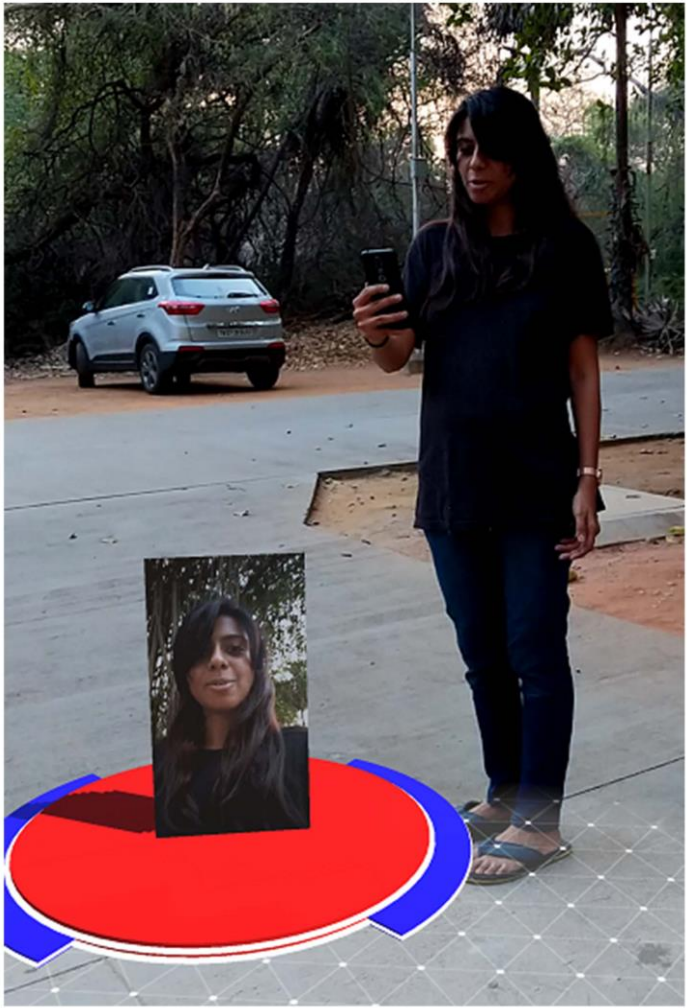
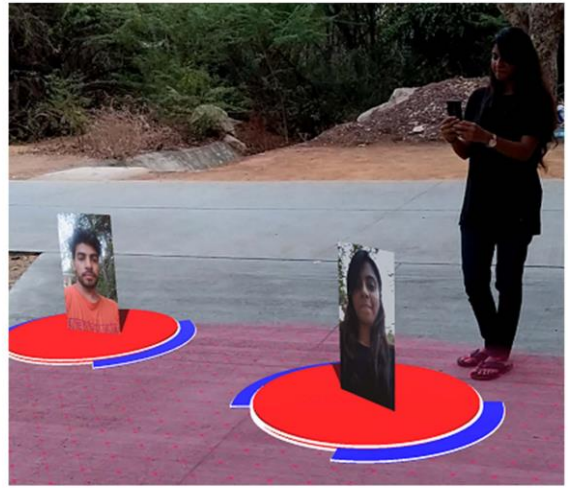
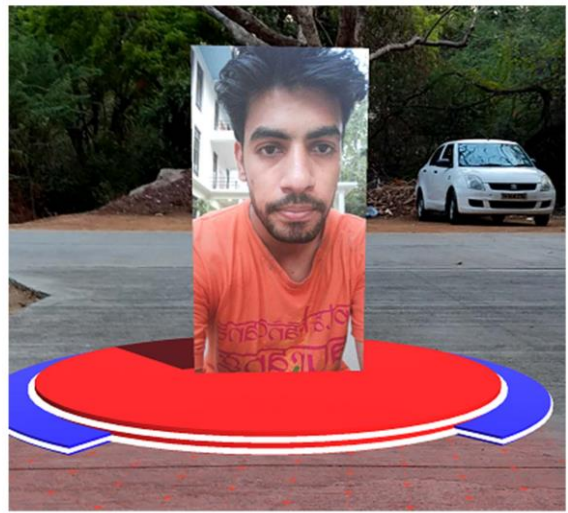
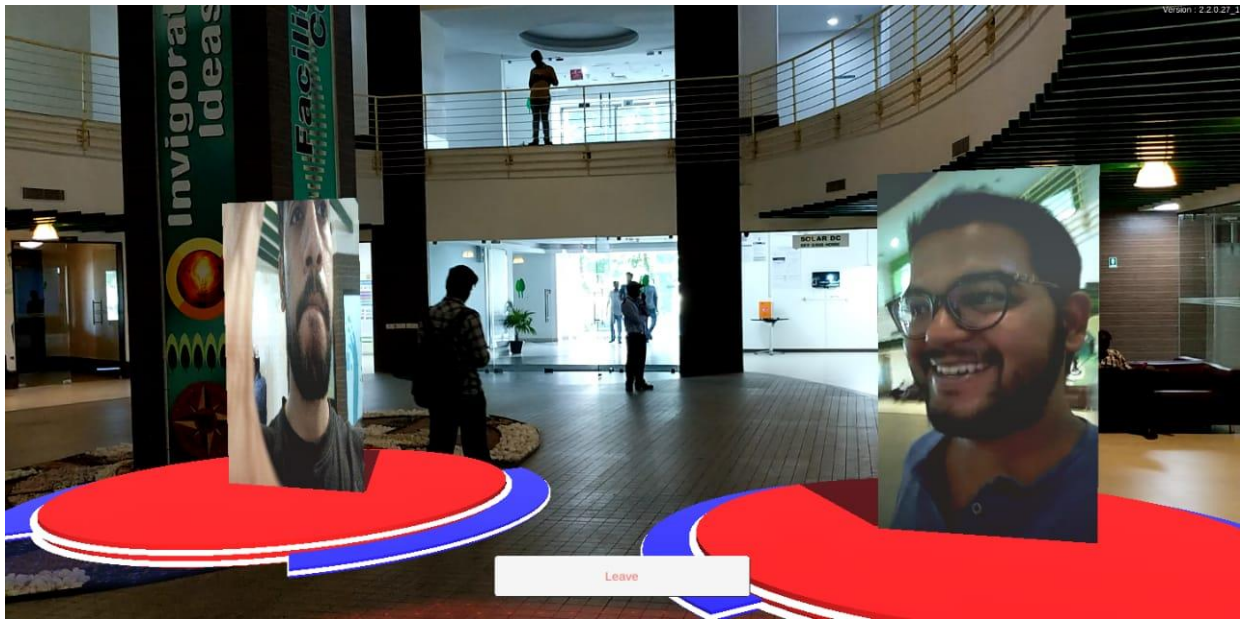

**Figure 5: More results of proposed AR video calling application; YouTube demo is available [13].**

## 5. APPLICATIONS

This concept is novel and may find route to its application in the VR/AR/MR and Mobile 3D TV research fields. Its major applications are foreseeable in communication. With the usage of Augmented Reality, the user gets a next level of immersive [6,8] experience while interacting. Augmented Reality helps in placing the participant in the real world, who is communicating virtually. This creates a trustable interaction and brings people closer without any requirement to strain oneself. Such forms of communication can help in personal as well as professional fronts like hang out and chat, collaborate and work, conduct virtual meetings, etc.

Apart from this, another important application of this technology can be seen in education, surveillance and medical industries. The extended version and study of this work will be very effective to provide real time instructions and training.

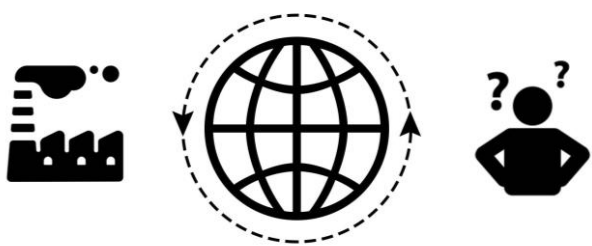

**Figure 6: Application of concept for industry**

For example, many on-site technical issues are unresolved due to various reasons. Time and monetary constraints may cause ineffective utilization of resources. This may also lead to serious immeasurable damages, if not fixed in time. Experts may be located in any nook and corner of the globe, but can provide instant advice and/or solution to the problem faced on the site using this tool. The proposed work can be very useful to give a clear view of the faulty equipment and the expert can analyze and rectify the problems right from their locations.

## 6. FUTURE WORK

Taking a step ahead, we are making Avatars, a replica of the participants interacting with each other in the real world. These avatars will depict the facial expressions, body language and movements. Using light field imaging techniques, these avatars will depict the characteristics of individuals. We will develop a technology that will enable the projection of 3D model of the people interacting with each other in real time. A single light-field camera will be used to capture light fields which will give us the capability to reconstruct a human face model based on the facial characteristics. This paves our path towards the concept of holoportation on Mobile devices using light fields with proposed tool and existing VR/AR HMDs [14].

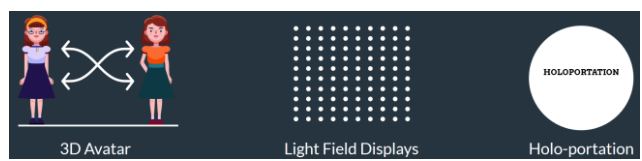

**Figure 7: Future Work**

## 7. ACKNOWLEDGMENT

The authors would like to thank Anusha Iyangar and students from "Computational Imaging and Display Lab" and "Advanced Geometric Computing Lab" at IIT Madras for their contribution in experiments and testing the tool.